\documentclass[letterpaper, 10 pt, conference,dvipsnames]{ieeeconf}  
\IEEEoverridecommandlockouts                              
\usepackage[utf8]{inputenc}
\usepackage[T1]{fontenc}
 \usepackage[export]{adjustbox} 
 \usepackage{diagbox}
\usepackage{colortbl}

\usepackage{color}
\usepackage{xcolor}
\usepackage{commath}
\usepackage{tabularx}
\usepackage{diagbox} 
\usepackage{cancel}
\usepackage{bm}

\usepackage{graphicx} 
\usepackage{amsmath} 
\usepackage{amssymb}  
\usepackage{cancel}
\usepackage{verbatim}
\usepackage{algorithm}
\usepackage{algpseudocode}
\usepackage{mathtools}
\usepackage{float}
\usepackage{tikz}
\usetikzlibrary{shapes.geometric}
\usetikzlibrary{decorations.pathmorphing}
\usetikzlibrary{arrows.meta}
\usepackage{comment}
\usepackage{subcaption}
\usepackage{hyperref}
\usepackage{siunitx}
\usepackage{soul}
\usepackage{booktabs}
\usepackage{multirow}
\usepackage{amsmath}
\usepackage{amssymb}
\usepackage{svg}

\usetikzlibrary{automata,positioning}

\tikzstyle{node}=[align=center]

\let\labelindent\relax
\usepackage[inline]{enumitem}

\include{h2t_def}

\title{\LARGE \bf

TOLEBI: Learning Fault-Tolerant Bipedal Locomotion via Online Status Estimation and Fallibility Rewards

}

 \author{Hokyun Lee$^{1}$, Woo-Jeong Baek$^{2*}$, Junhyeok Cha$^{1}$, and Jaeheung Park$^{1,3*}$
\thanks{This work was supported by the Technology Innovation Program (RS-2025-25453780, Development of a National Humanoid AI Robot Foundation Model for Multi‑Task Applications) funded By the Ministry of Trade, Industry, and Resources (MOTIR, Korea).}
\thanks{$^{1}$Hokyun Lee, Junhyeok Cha and Jaehueng Park are with the Department of Intelligence and Information, Graduate School of Convergence Science and Technology, Seoul National University, Seoul 08826, Republic of Korea. {\tt\small [hkleetony, threeman1, park73]@snu.ac.kr}}%
\thanks{$^{2}$Woo-Jeong Baek is with the Artificial Intelligence Institute (AIIS), Seoul National University, Republic of Korea. {\tt\small wjbaek@snu.ac.kr}}%
\thanks{$^{3}$Jaeheung Park is also with Advanced Institutes of Convergence
Technology (AICT), Suwon 16229, Republic of Korea and with ASRI, AIIS, Seoul National University, Seoul 08826, Republic of Korea.}
\thanks{*Corresponding authors: Jaeheung Park and Woo-Jeong Baek}%
 }

\makeatletter
\def\namedlabel#1#2{\begingroup
    #2%
    \def\@currentlabel{#2}%
    \phantomsection\label{#1}\endgroup
}
\makeatother
\begin{document}

\maketitle

\begin{abstract}
With the growing employment of learning algorithms in robotic applications, research on reinforcement learning for bipedal locomotion has become a central topic for humanoid robotics. While recently published contributions achieve high success rates in locomotion tasks, scarce attention has been devoted to the development of methods that enable to handle hardware faults that may occur during the locomotion process. However, in real-world settings, environmental disturbances or sudden occurrences of hardware faults might yield severe consequences. To address these issues, this paper presents \textit{TOLEBI: A fault-tolerant learning framework for bipedal locomotion} that handles faults on the robot during operation. Specifically, joint locking, power loss and external disturbances are injected in simulation to learn fault-tolerant locomotion strategies. In addition to transferring the learned policy to the real robot via sim-to-real transfer, an online joint status estimator incorporated. This module enables to classify joint conditions by referring to the actual observations at runtime under real-world conditions. The validation experiments conducted both in real-world and simulation with the humanoid robot TOCABI highlight the applicability of the proposed approach. To our knowledge, this work provides the first learning-based fault-tolerant framework for bipedal locomotion, thereby fostering the development of efficient learning methods in this field.
\end{abstract}

\section{Introduction}
\noindent Dealing with unexpected system failures during operation is one crucial issue in real-world robotic applications. While extensively studied across robotics subdomains, the derivation of suitable techniques that enable to avoid the negative consequences of undesired faults has become increasingly challenging with the growing employment of learning algorithms. Indeed, the performance and flexibility of robot applications has been improved significantly with advances in the learning domain. For example, efforts for deriving complex robot control algorithms, like bipedal locomotion for humanoid robots or manipulation tasks can be efficiently reduced via reinforcement learning \cite{zhang2024whole, zhang2024achieving}.
However, one drawback of learning methods is their black-box character that makes the prediction regarding unseen data difficult \cite{schelter2020learning, redyuk2019learning}. As a consequence, compromises that reduce the flexibility and retain robot safety are required to facilitate the application of according systems under real-world conditions. In the last decade, bipedal locomotion has gained increased attention \cite{tong2024advancements} in the humanoid robotics domain. However, controllers that handle hardware failures in bipedal locomotion are missing in current robotics literature. In contrast to quadruped robots, reduced functionalities on one leg can significantly degrade the system performance for biped robots. Sudden occurrences of faults on one leg can result in situations where the robot loses its balance and falls.
Particularly, these events can occur in an unexpected manner at runtime due to environmental disturbances or undesired force perturbations \cite{winfield2006safety, huber2000hybrid}. Therefore, the derivation of methods that can deal with faults inherently is essential. One critical challenge here lies in retaining the benefits of reinforcement learning while ensuring that faults and their consequences can be mitigated despite the black-box character of learning algorithms.
\begin{figure}[t]
  \centering
  \includegraphics[width=\linewidth]{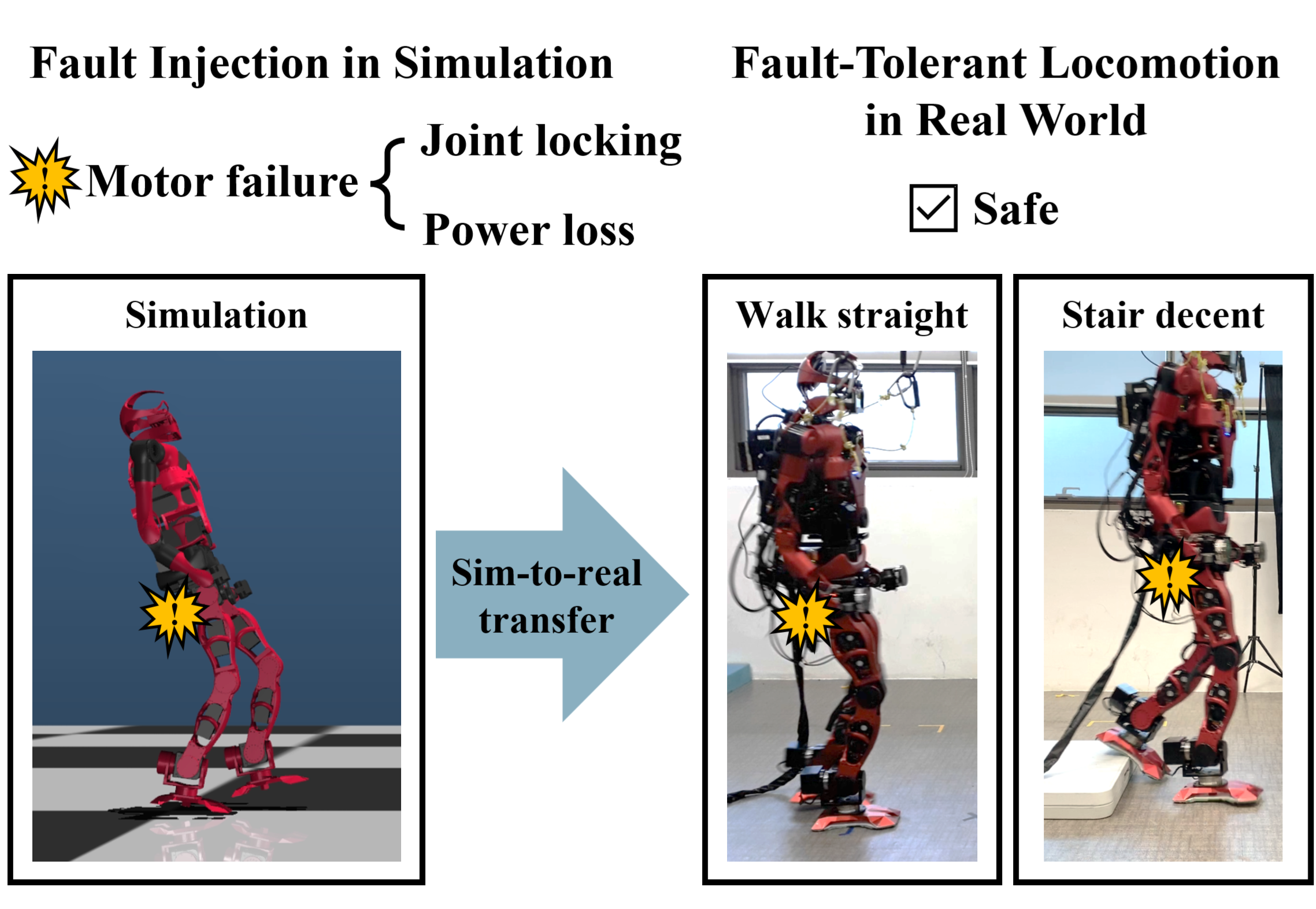}
  \caption{\textbf{TOLEBI, A framework for learning fault-tolerant bipedal locomotion.} Motor failures are injected during training in simulation to learn a fault-tolerant locomotion policy, and the policy is transferred to the real humanoid robot.}
  
  \label{fig:overview}
\end{figure}
Recent works that present fault-tolerant strategies for robot locomotion have been designed for quadruped robot systems, where the fault behavior is assumed for one of four legs. While promising performances have been achieved, these approaches cannot be directly applied to bipedal locomotion. Furthermore, model-based methods for quadruped systems that handle faults by incorporating them in the control explicitly like \cite{cui2022fault} exist. Here, the faults are modeled and incorporated manually into the controller. While their performance regarding the fault detection and failure reduction has been validated, their applicability is limited to the situations considered in the models. Therefore, model-based approaches struggle to deal with unseen situations.
In order to derive a locomotion framework for bipedal robot locomotion that exploits the benefits of learning and can handle faults, this paper presents \textbf{TOLEBI} (a faul\textbf{T}-t\textbf{O}lerant \textbf{L}earning fram\textbf{E}work for \textbf{B}ipedal locomot\textbf{I}on) based on reinforcement learning with the humanoid robot TOCABI \cite{schwartz2022design}, as shown in Fig.~\ref{fig:overview}. In particular, TOLEBI leverages phase modulation actions and the proposed fallibility rewards to derive a control policy that can deal with faults while transferring the policy successfully to the real robot. 
To be specific, the reward is designed with respect to the foot force aiming that the contact force between the foot and the floor is reduced. Scientifically, the novelty of the TOLEBI lies in training a joint status estimator online on the basis of a curriculum of motor failure simulations. The online estimation of the current joint status is considered as the observation, which enables to achieve higher robustness. The performance of TOLEBI is evaluated on two real-world locomotion scenarios: Walking on a plain floor and descending stairs.
Therefore, this manuscript presents the first work in the robotics domain that presents a learning-based fault-tolerant bipedal locomotion approach for real-world environments. 
The remainder of this paper is structured as follows: Section~\ref{sec:sota} introduces state-of-the-art literature in the field of learning for bipedal locomotion, thereby highlighting the scientific novelties of the TOLEBI. After summarizing the preliminaries in Section~\ref{sec:preliminaries}, TOLEBI is derived in Section~\ref{sec:method} by specifying the reinforcement learning algorithm with its rewards. In addition, the policy learning approach and the sim-to-real transfer are described in detail. The results of the validation experiments are presented in Section~\ref{sec:experiments}. Finally, Section~\ref{sec:conclusion} summarizes the scientific findings and suggests directions for future research. 

\section{Related Work} \label{sec:sota}
\subsection{Reinforcement Learning for Bipedal Locomotion}
\noindent Robot locomotion control has long been a central research topic in robotics, and recent advances in deep reinforcement learning (DRL) have established it as a key methodology for developing robust controllers in complex and dynamic environments. DRL has been successfully applied to a wide range of tasks including fall recovery control for quadruped robots \cite{lee2019robust}, locomotion control conditioned on parameterized gait and task inputs for bipedal robots \cite{li2021reinforcement, li2025reinforcement}, and improving controller generalization through diverse behavioral training \cite{margolis2023walk}. More recently, real-world deployment of DRL-based controllers has enabled successful locomotion on humanoid robots \cite{radosavovic2024real}. These advances have been largely facilitated by high-performance simulation environments such as Isaac Gym \cite{makoviychuk2021isaac} that enable to process large-scale parallel data collection with high-fidelity physics, thereby bridging the sim-to-real gap and allowing the development of versatile, dynamic, and transferable locomotion policies like presented in the contributions \cite{kim2023torque, kim2024bridging}.

\subsection{Fault-tolerant Locomotion}
\noindent When robots operate in real-world environments, unexpected hardware failures such as actuator malfunctions can severely degrade performance or lead to complete loss of mobility. Fault-tolerant locomotion aims to maintain stability and accomplish mission objectives even under such fault conditions, significantly enhancing the reliability and practicality of legged robots. These can be split in model-based and learning-based approaches: 

\subsubsection{Model-based Methods}
Early research on fault-tolerant control mainly relies on model-based methods, where accurate kinematic and dynamic models were used to design alternative gaits for failed actuators in quadruped \cite{yang2006kinematic, pana2008fault}, and hexapod robots \cite{asif2012improving, yang1998fault}. Approaches such as posture optimization and whole-body control were introduced to handle specific fault scenarios in quadruped \cite{cui2022fault}. However, these methods often required extensive manual modeling, showed limited scalability in unstructured environments, and faced difficulties in dealing with perturbations or failures.

\subsubsection{Learning-based Methods}
To address these limitations, learning approaches have recently gained attention. As mentioned above, Reinforcement learning (RL) enables robots to learn robust control policies through interaction with diverse failure scenarios in simulation, eliminating the need for explicit fault modeling. In contrast, meta-learning techniques allow rapid adaptation to changing dynamics \cite{anne2021meta}, while curriculum learning strategies gradually increase fault severity during training to improve robustness \cite{okamoto2021reinforcement}. Recent works have proposed RL-based frameworks for quadrupedal robots that incorporate fault detection, recovery, and locomotion adaptation under degraded actuator conditions \cite{luo2023ft, wu2023adaptive, liu2025fault}. Methods such as random joint masking \cite{kim2024learning} and multi-task learning \cite{hou2024multi} further enhance generalization across different failure modes. 

\noindent Despite these advances, the majority of existing methods focus on quadrupedal robots, where stability margins are inherently larger. In contrast, bipedal locomotion presents additional challenges due to reduced stability and higher risk of catastrophic falls under motor failures on one leg. This motivates the development of fault-tolerant learning frameworks specifically designed for humanoid robots operating under unexpected joint locking or power loss events. 
Therefore, the scientific contributions of this manuscript can be summarized as follows:
\begin{enumerate}
    \item First, a fault-tolerant learning framework for bipedal locomotion \textbf{TOLEBI} is proposed that incorporates a curriculum learning approach and motor failure simulations to facilitate sim-to-real transfer.
    \item Second, we integrate the online joint status estimator concurrently trained with the policy to infer joint status without additional training phases into TOLEBI.
    \item Third, the fallibility reward is designed for TOLEBI under motor failure while preserving nominal locomotion.
\end{enumerate}

\begin{figure*}[ht]
    \centering
    \includegraphics[width=\textwidth]{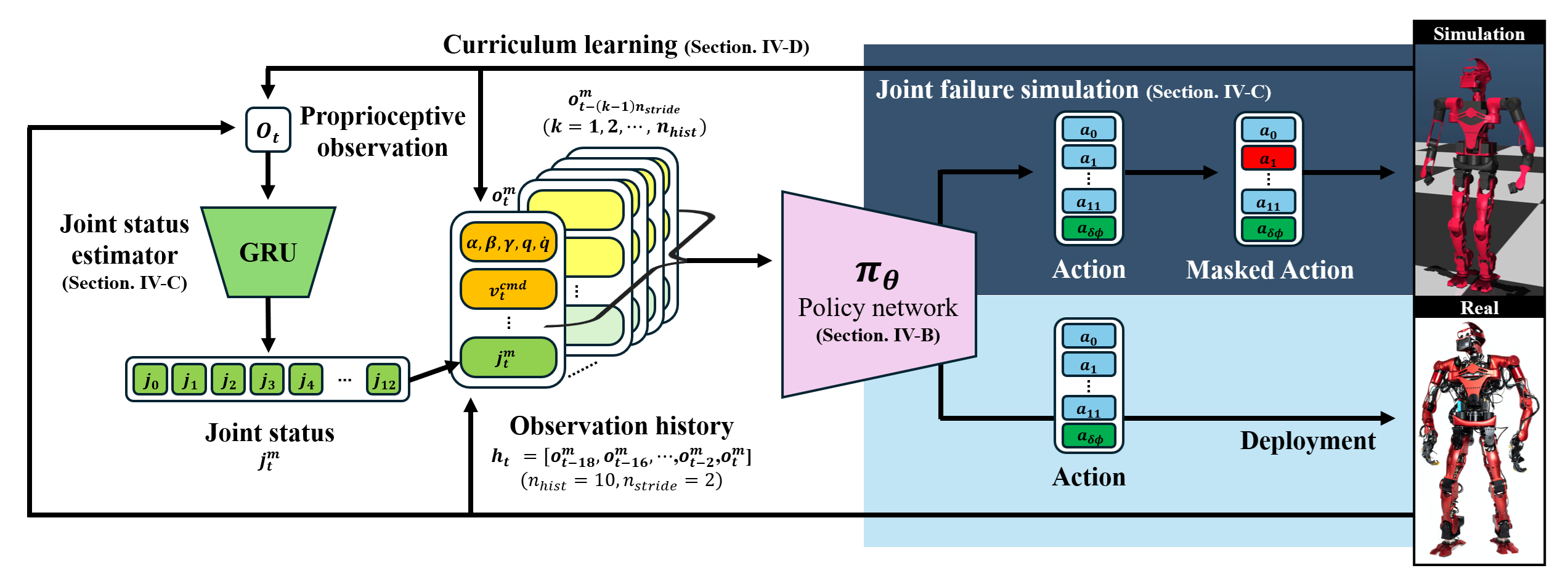}
    \caption{\textbf{Schematic description of the framework TOLEBI (faulT-tOlerant Learning framEwork for Bipedal locomotIon).} A joint status estimator processes proprioceptive observations to infer joint status, storing the results in the observation history for policy training. During simulation, motor failure scenarios mask the corresponding actions, enabling robust fault-tolerant policy learning. The trained policy is deployed on the real humanoid robot for fault-tolerant locomotion.}
    \label{fig:framework}
\end{figure*}

\section{Preliminaries} \label{sec:preliminaries}
\subsection{Reinforcement Learning}
\noindent We formulate the bipedal locomotion control problem under motor failure as a Markov Decision Process (MDP). The MDP defined by a tuple ($\mathcal{S},\mathcal{A},\mathcal{P},\mathcal{R},\gamma$) that is composed of the state space $\mathcal{S}$, the action space $\mathcal{A}$, a transition probability function $\mathcal{P}$, a reward function $\mathcal{R}$, and a discount factor $\gamma$. The agent aims to learn a policy $\pi_\theta(a|s)$ by interacting with the environment to maximize the expected discounted return $J(\pi_\theta)$ defined as the cumulative discounted reward over a finite-horizon $T$:
\begin{equation}
    J(\pi_\theta) = \mathbb{E}_{\tau \sim p(\tau|\pi_\theta)} \left[\sum_{t=0}^{T-1} \gamma^t r_t\right]
\end{equation}
where $\tau$ denotes a trajectory generated under the policy $\pi_\theta$.

\subsection{Motor Failures}
\noindent Motor failures in robots primarily arise from two sources. External disturbances can mechanically constrain the actuators while internal faults in electrical or control subsystems might disrupt power transmission or motor command signals. Both failure types can lead to significant performance degradation or even a complete loss of mobility during bipedal locomotion.
This paper focuses on two failure scenarios: Joint locking and power loss. Here, \textbf{joint locking} occurs when actuators become stuck and prevent any motion in the affected joints. In contrast, \textbf{power loss} arises when joints remain physically free but cannot generate torque. Both failures severely compromise the robot's balance, hinder the ability to track commanded velocities, and disrupt stable locomotion.

\section{TOLEBI - A Fault-Tolerant Learning Framework for Bipedal Locomotion} \label{sec:method}
\noindent This section introduces TOLEBI: A reinforcement learning framework for fault-tolerant bipedal locomotion under unexpected motor failures. The proposed approach integrates motor failure simulation with joint masking, joint status estimation, curriculum learning and an action structure that incorporates phase modulation for adaptive gait timing. The entire control strategy is trained using Proximal Policy Optimization (PPO) \cite{schulman2017proximal} in Issac Gym and deployed on the real robot to ensure robustness at operation.



\subsection{Overview}
\noindent Our proposed framework TOLEBI shown in Fig.~\ref{fig:framework} integrates joint status estimation, observation history modeling, and curriculum learning for fault-tolerant bipedal locomotion under motor failures. At each time step, the observation includes proprioceptive observation, command velocities, and the estimated joint status with a history buffer that stores the latest $n_{history}=10$ entries with a stride of $n_{stride}=2$ between them. This approach allows the policy to leverage sequential information without significant computational overhead.
Motor failures are simulated by masking joint torque commands in Section~\ref{sec:failure_simulation} enabling the policy to learn control under partial actuation. The joint status estimator appends inferred joint status to the observation history during curriculum learning in Section~\ref{sec:curriculum}.



\begin{table*}[ht!]
\centering
\caption{REWARD FUNCTIONS AND SCALES AT EACH CURRICULUM LEARNING PHASE}
\label{tab:reward_phases}
\begin{tabular}{ll l c c}
\toprule
\textbf{Category} & \textbf{Reward} & \textbf{Equation} ($r_i$) & \textbf{Scale in} & \textbf{Scale in} \\
&&& \textbf{nominal phase} ($w_i$)& \textbf{fault phase} ($w_i$)\\
\midrule
\multirow{3}{*}{Task Rewards} 
& Linear velocity tracking & $\exp(-\frac{1}{0.45^2} \| \mathbf{v}_{xy}^{cmd} - \mathbf{v}_{xy} \|_2^2)$ & 0.4 & 0.4 \\
& Angular velocity tracking & $\exp(-\frac{1}{0.35^2} \| \mathbf{w}_{z}^{cmd} - \mathbf{w}_{z} \|^2)$ & 0.2 & 0.2 \\
& Foot contact & $\mathbf{1}_{\text{DSP/RSSP/LSSP\,sync}}$ & 0.2 & 0.2 \\
\midrule
\multirow{8}{*}{Regulation Terms} 
& Body orientation & $\exp(-500.0 ({roll}^2+{pitch}^2))$ & 0.3 & 0.3 \\
& Joint torque & $\exp(-\frac{1}{100.0}\| \boldsymbol{\tau} \|)$ & 0.05 & 0.05 \\
& Joint velocity & $\exp(-\frac{1}{100.0}\| \dot{\boldsymbol{q}} \|)$ & 0.05 & 0.05 \\
& Joint acceleration & $\exp(-\frac{1}{0.05}\| \ddot{\boldsymbol{q}} \|)$ & 0.05 & 0.05 \\
& Feet contact force & $\exp(
    -\tfrac{1}{140.0} 
    \sum\limits_{i \in \{L,R\}} 
    \lVert \mathrm{ReLU}(\mathbf{F}_z^i - 1.4W) \rVert
)$ & 0.1 & 0.1 \\
& Torque difference & $\exp(-\frac{1}{1.20^2}\| \mathbf{\tau}_t-\mathbf{\tau}_{t-1} \|)$ & 0.7 & 0.7 \\
& Contact force difference & $\exp(-\frac{1}{100.0}\sum\limits_{i \in \{L,R\}}\| \mathbf{F}_{z,t}^i-\mathbf{F}_{z,t-1}^i \|)$ & 0.2 & 0.2 \\
\midrule
\multirow{3.5}{*}{Fallibility Rewards}
& Trajectory mimicking & $\exp(-\frac{1}{0.5} \| \mathbf{q}^{ref} - \mathbf{q}\|^2)$ & \textbf{0.35} & \textbf{0.35} \\
& Contact force tracking & $\exp(-\frac{1}{10.0}\sum\limits_{i \in \{L,R\}}\| \mathbf{F}_{z}^{i,ref} - \mathbf{F}_{z}^i \|)$ & 0.0 & \textbf{0.3} \\
& Termination penalty & $\mathbf{1}_{\text{terminate}}$ & 0.0 & \textbf{-100.0} \\
\bottomrule
\end{tabular}
\end{table*}

\subsection{Reinforcement Learning for Biped Locomotion}
\subsubsection{State Space}
The state $\mathcal{S} \in \mathbb{R}^{51}$ consists of the robot's base orientation in Euler angles $\Theta_t =(\alpha_t, \beta_t, \gamma_t) \in \mathbb{R}^3$, joint positions $q_t \in \mathbb{R}^{12}$ in the legs, joint velocities $\dot{q}_t \in \mathbb{R}^{12}$, and the walking phase information encoded as the sine and cosine of its phase $\Phi_t = (\mathcal\sin(2\pi \phi_t), \cos(2\pi \phi_t)) \in \mathbb{R}^2$. The phase variable $\phi_t$ evolves cyclically from 0 to 1. 
The command velocity $v^{cmd}_t = (v^x_t, v^y_t, \omega_t^z) \in \mathbb{R}^3$, and base velocity $v^{base}_t \in \mathbb{R}^6$.
The state also incorporates a joint status vector $j^m_t \in \mathbb{R}^{13}$ inferred online by a GRU-based joint status estimator during training. Formally, the state is expressed as:
\begin{equation}
    s_t = \{\Theta_t, q_t, \dot{q}_t, \Phi_t, v^{cmd}_t, v^{base}_t, j^m_t \}
\end{equation}
In the joint status vector, the first dimension indicates whether the entire system is in a healthy state, while the remaining 12 dimensions represent the operational status of individual motors.

\subsubsection{Action Space}
The action space $\mathcal{A} \in \mathbb{R}^{13}$ comprises 12 joint torque commands and an additional action for modulating the phase $\phi_t$. At each timestep, the policy outputs the mean $\mu_\theta$ of a Gaussian distribution $\mathcal{N}(\mu_\theta, \sigma^2)$, from which actions are sampled. The standard deviation $\sigma$ is predetermined according to the torque limits of individual joints for exploration. The action space also includes a phase modulation action $a_{\delta \phi,t}$, which plays a critical role in adapting locomotion timing under motor failures. To enable rapid adaptation when actuators malfunction, the phase modulation action adjusts the motion period and timing as follows:
\begin{equation}
\phi_{t+1} = \left(\phi_t + \frac{\Delta t}{T_{ref}} + a_{\delta \phi,t}\right) \bmod 1.0
\end{equation}
By directly influencing the gait timing, $a_{\delta \phi,t}$ allows the policy to modify the motion cycle and maintain stability under unexpected motor failures.

\subsubsection{Reward Function}
The reward function is designed to ensure stable and fault-tolerant bipedal locomotion while promoting energy efficiency, smooth control, and sim-to-real transfer. At each timestep, the total reward is defined as the weighted sum of multiple components as follows:
\begin{equation}
r_{total} = r_{task} + r_{regulation} + r_{fall}
\end{equation}

\noindent \textbf{Task Rewards.} Task rewards encourage the robot to track commanded linear and angular velocities and maintain proper foot contacts. The foot contact reward $r_c$, is given for matching the gait cycle, which is divided into Double Support Phase (DSP), Right Single Support Phase (RSSP), and Left Single Support Phase (LSSP):
\begin{equation}
r_{task} = w_{v,xy} r_{v,xy} + w_{w,z} r_{w,z} + w_c r_c
\end{equation}

\noindent \textbf{Regulation Terms.}
We introduce the regulation terms to improve motion smoothness, energy efficiency, and physical feasibility during locomotion. These terms penalize sudden or excessive movements in body posture, joint dynamics, and interaction forces. The total regulation reward is given by
\begin{equation}
\label{eq:regulation_reward}
\begin{split}
    r_{\text{regulation}} = w_o r_o + w_\tau r_\tau + w_v r_v + w_a r_a + w_f r_f \\ +w_{\Delta\tau} r_{\Delta\tau} + w_{\Delta f} r_{\Delta f}, 
\end{split}
\end{equation}
where each weight $w$ balances the contribution of the respective term.



\noindent \textbf{Fallibility Rewards.} Finally, the proposed fallibility reward is a composite function designed to maintain stable locomotion under motor failures. The first is a trajectory mimic reward $r_q$, encourages tracking of nominal joint trajectory $q^{ref}$ even under faulty conditions. Alternative strategies for quadruped locomotion—such as removing the reward \cite{kim2024learning} or adding rewards to lift the leg with the failed motor \cite{wang2024learning}—turned out to be unsuitable for bipedal locomotion in our work. Removing the reward caused the policy to adopt an overly stable, crouching gait that lost its natural walking style and was not transferable to the real robot. A foot-lifting strategy is similarly impractical, as it requires the robot to hop on one leg. The second term, a force reference reward $r_{f,ref}$ mitigates large impacts caused by early foot contacts under fault conditions by encouraging adherence to the reference foot contact force $F_z^{ref}$. Finally, a termination penalty $r_T$ imposes a strong penalty when the episode ends due to falling or self collisions:

\begin{equation}
r_{fall} = w_q r_q + w_{f,ref} r_{f,ref} + w_T r_T
\end{equation}
where the definitions of the reward terms $r_i$ and the weights $w_i$ that balance their contributions are listed in Table~\ref{tab:reward_phases}.




\subsection{Motor Failure Simulation and Status Estimation}
\label{sec:failure_simulation}
\subsubsection{Failure Simulation}
\label{subsec:failure_simulation}
We simulate two types of motor failures: Joint locking and power loss by masking the action commands during training. For each training iteration, 90\% of the environments are randomly assigned to a fault condition. In this subset, the failure type is sampled with a 50\% probability for each case. Next, a joint index $j \in \{0, 1, \dots, 11\}$ is uniformly selected to mask its corresponding action.


\textbf{Joint Locking.} The joint torque is computed via predefined proportional gains $K_p$ and derivative gains $K_d$ to hold the current joint position $q_j^0$ fixed at the moment of failure.

\textbf{Power Loss.} The commanded torque is set to zero to effectively disable actuation. Formally, the masked torque command $\tau_j$ for joint $j$ is given by
\begin{equation}
\label{eq:torque_masking}
\tau_j = 
\begin{cases} 
    K_p(q_j^0 - q_j) - K_d \dot{q}_j, & \text{if joint locking} \\
    0, & \text{if power loss} \\
    \tau_j, & \text{otherwise}
\end{cases}
\end{equation}
Since our policy directly outputs the joint torque commands $\tau_j$ as actions, this masking strategy imposes realistic failure conditions and enables the policy to learn robust locomotion behaviors under partial actuation.

\subsubsection{Status Estimation}
A joint status estimator based on a single-layer GRU with a hidden size of 128 and a learning rate of $10^{-4}$ is trained online to infer joint status from proprioceptive inputs. A joint status is classified as faulty (status = 1) when the estimator output, produced by a sigmoid activation in the range $[0, 1]$, exceeds a threshold of $0.7$. The estimator does not distinguish between joint locking and power loss, considering both as fault status. 
The estimator is optimized using the BCE loss between the predicted probabilities and the actual masking indices, with the threshold applied only at the decision stage rather than during training. The online learning scheme updates the estimator continuously in parallel with policy training, allowing it to gradually adapt to changing joint status. The estimated joint status vector is then appended to the observation space so that the policy can adjust its control commands based on real-time estimates of motor health.

\begin{algorithm}[t]
\caption{Curriculum Policy Learning}
\label{alg:curriculum}
\begin{algorithmic}[1]
\State Initialize policy $\pi_\theta$ with PPO
\For{$epoch = 1, 2, \dots, N$}
    \State Collect rollouts with current policy $\pi_\theta$
    \State Compute average episode length $L_k$
    \If{$L_k > 20\,\mathrm{s}$ \textbf{and} joint masking not enabled}
        \State Enable motor failure simulation (Sec.~\ref{sec:failure_simulation})
    \EndIf
    \If{$L_k > 24\,\mathrm{s}$ \textbf{and} perturbations not enabled}
        \State Enable push perturbations (Sec.~\ref{sec:sim2real})
    \EndIf
    \State Update policy $\pi_\theta$ using PPO
\EndFor
\end{algorithmic}
\end{algorithm}

\subsection{Curriculum Learning}
\label{sec:curriculum}

\noindent Curriculum learning plays an important role in stabilizing training for complex tasks like fault-tolerant humanoid locomotion. Rather than exposing the policy to all challenging conditions from the beginning, TOLEBI gradually increases task complexity based on the agent’s performance, allowing the policy to acquire fundamental locomotion skills before adapting to motor failures and external disturbances.
\noindent As outlined in Algorithm~\ref{alg:curriculum} the training starts with nominal locomotion under ideal conditions without failures. This approach is taken because our preliminary experiments showed that premature exposure to faults led to unstable training and degraded performance. When the average episode length exceeds 20 seconds, motor failure simulation is introduced following the masking strategy in Section~\ref{sec:failure_simulation}. The fallibility rewards are adjusted according to Table~\ref{tab:reward_phases}. where premature exposure to faults led to unstable training and degraded performance.
\noindent Subsequently, when the average episode length under motor failures surpasses 24 seconds, push perturbations are applied to the robot's base to improve robustness for sim-to-real transfer. This progressive curriculum provides a structured path from nominal locomotion to fault-tolerant and disturbance-resilient behaviors.
\newcommand{\rot}[1]{\rotatebox{90}{#1}}


\subsection{Sim-to-Real Transfer}
\label{sec:sim2real}

\noindent To bridge the gap between simulation and real-world deployment, both domain randomization and dynamics randomization techniques \cite{margolis2023walk, kim2024bridging, tobin2017domain} are employed during training, as summarized in Appendix, Table~\ref{tab:randomization_params_compact}.

\subsubsection{Domain Randomization} Environmental and sensory variations are introduced to improve the policy’s robustness against real-world uncertainties. Commanded velocities are randomized, horizontal push perturbations are randomly applied, and noise is injected into base velocity measurements to account for sensing and actuation inaccuracies.

\subsubsection{Dynamics Randomization} Physical parameters of the robot, including motor constants, base mass, joint damping, inertia, and actuation delays, are randomized to bridge the modeling gap between simulation and reality. These randomization strategies enhance the policy’s generalization capability, enabling robust performance when deployed on the real humanoid robot.

\begin{table}[t!]
    \centering
    \caption{\textbf{Success rate across failure scenarios in  simulation.}
Locomotion success rates are shown for the baseline, our method with joint masking and status estimation, and the full approach with curriculum learning with the fallibility reward $r_{fall}$. \textbf{Bold} indicates the best performance while \underline{underlined} values denote no successful trials were recorded.}
    \label{tab:failure_scenario_comparison}
    \setlength{\tabcolsep}{4pt} 
    \begin{tabular}{llccc}
        \toprule
        \multicolumn{2}{c}{\textbf{Scenarios}} & \textbf{Baseline} \cite{kim2023torque} & \textbf{\begin{tabular}[c]{@{}c@{}}$+$ joint mask. \\ and status est.\end{tabular}} & \textbf{\begin{tabular}[c]{@{}c@{}}$+$ curriculum \\ and $r_{fall}$ (Ours)\end{tabular}} \\
        \midrule
       & {Healthy} & \textbf{0.9893} & 0.5237 & 0.9624 \\
        \midrule
        \multirow{7}{*}{\rot{\textbf{Joint locking}}} 
        & hip yaw     & 0.2378 & 0.7063 & 0.9194 \\
        & hip roll    & \underline{0.0000} & 0.6956 & 0.7974 \\
        & hip pitch   & \underline{0.0000} & 0.3406 & 0.7073 \\
        & knee pitch  & 0.1462 & 0.4856 & 0.8130 \\
        & ankle pitch & \underline{0.0000} & 0.2683 & 0.6440 \\
        & ankle roll  & 0.1150 & 0.5398 & 0.9951 \\
        \cmidrule(lr){2-5}
        & \textbf{Average} & 0.0832 & 0.5060 & \textbf{0.8127} \\
        \midrule
        \multirow{7}{*}{\rot{\textbf{Power Loss}}}
        & hip yaw     & 0.6187 & 0.8989 & 0.9753 \\
        & hip roll    & \underline{0.0000} & \underline{0.0000} & \underline{0.0000} \\
        & hip pitch   & \underline{0.0000} & 0.5647 & 0.5784 \\
        & knee pitch  & \underline{0.0000} & \underline{0.0000} & \underline{0.0000} \\
        & ankle pitch & \underline{0.0000} & 0.5205 & 0.6189 \\
        & ankle roll  & 0.7173 & 0.7949 & 0.9873 \\
        \cmidrule(lr){2-5}
        & \textbf{Average} & 0.2227 & 0.4632 & \textbf{0.5267} \\
        \bottomrule
    \end{tabular}
\end{table}

\section{Experimental Results} \label{sec:experiments}

\subsection{Experimental Setup}
\noindent The proposed policy is trained on the PPO algorithm with both the actor and critic networks modeled as two hidden layer MLPs consisting of 256 ReLU units per layer. The training is performed in the Isaac Gym simulator with 4096 parallel environments at a simulation frequency of 500 Hz and a control rate of 250 Hz. The maximum episode is set to 32 seconds. At each training iteration, a batch of 16,384 samples is collected, and the policy parameters are updated with a mini-batch size of 128 and a learning rate that linearly decays from $10^{-5}$ to $3 \times 10^{-6}$.


\subsection{Performance of Simulated Failure Scenarios}
\noindent Table~\ref{tab:failure_scenario_comparison} reports the locomotion success rates across different motor failure scenarios in the Isaac Gym simulation. These rates are calculated by counting the number of successful episodes in 4096 parallel environments, where a successful episode is defined as one that lasts for at least 20 seconds. We compare three methods: The baseline policy \cite{kim2023torque}, the policy trained with joint masking and status estimation, and our complete approach with the curriculum learning and the proposed fallibility reward.
The results demonstrate the enhanced performance of our method. Under joint locking conditions, the baseline policy fails to maintain stable locomotion in most cases with multiple scenarios showing zero success rates. Incorporating joint masking and status estimation significantly improves the performance. The success rate remains limited in challenging cases like knee pitch and ankle pitch power loss failures.
Our full approach achieves the highest average success rates under both joint locking (81.27\%) and power loss (52.67\%) conditions. The staged curriculum learning allows the policy to first acquire nominal locomotion skills prior to progressively handling motor failures and external perturbations while the fallibility reward encourages the robust recovery during fault-tolerance actuation. These results confirm that both components are essential for fault-tolerant locomotion across diverse failure scenarios.

\begin{table}[t!]
    \centering
    \caption{\textbf{Ablation study results for velocity tracking.} \textbf{Bold} indicated the best performance, while \underline{underlined} values denote the worst performance among the compared methods.}
    \label{tab:ablation_study_with_mbe_compact}
    \setlength{\tabcolsep}{5pt}
    \begin{tabular}{llS[table-format=1.4]S[table-format=+-1.4]}
        \toprule
        \multicolumn{2}{l}{} & {\textbf{Lin Vel [m/s]}} & {\textbf{Ang Vel [rad/s]}} \\
        & {\textbf{Conditions}} & {RMSE} & {RMSE} \\
        & {} & {\textit{(MBE)}} & {\textit{(MBE)}} \\
        \midrule
        \multirow{8}{*}{\rot{\textbf{Ablation}}} 
        & \multirow{2}{*}{w/o Joint status observation} & 0.1795 & \underline{0.2074} \\
        & & \textit{(-0.0512)} & \underline{\textit{(-0.0831)}} \\ 
        \cmidrule(lr){2-4}
        & \multirow{2}{*}{w/o Fallibility rewards} & 0.1529 & 0.1911 \\
        & & \textit{(-0.0450)} & \textit{(-0.0799)} \\
        \cmidrule(lr){2-4}
        & \multirow{2}{*}{w/o Phase modulation} & \underline{0.2190} & 0.1499 \\
        & & \underline{\textit{(-0.1585)}} & \textit{(-0.0614)} \\ 
        \cmidrule(lr){2-4}
        & \multirow{2}{*}{w/o Curriculum learning} & 0.2017 & 0.1320 \\
        & & \textit{(-0.0411)} & \textit{(-0.0559)} \\ 
        \midrule
        \multirow{2}{*}{} & \multirow{2}{*}{\textbf{Ours (TOLEBI)}} & \textbf{0.0833} & \textbf{0.1110} \\
        & & \textit{\textbf{(-0.0346)}} & \textit{\textbf{(-0.0016)}} \\ 
        \bottomrule
    \end{tabular}
\end{table}

\begin{figure}[hb!!!]
    \centering
    \includegraphics[width=\linewidth]{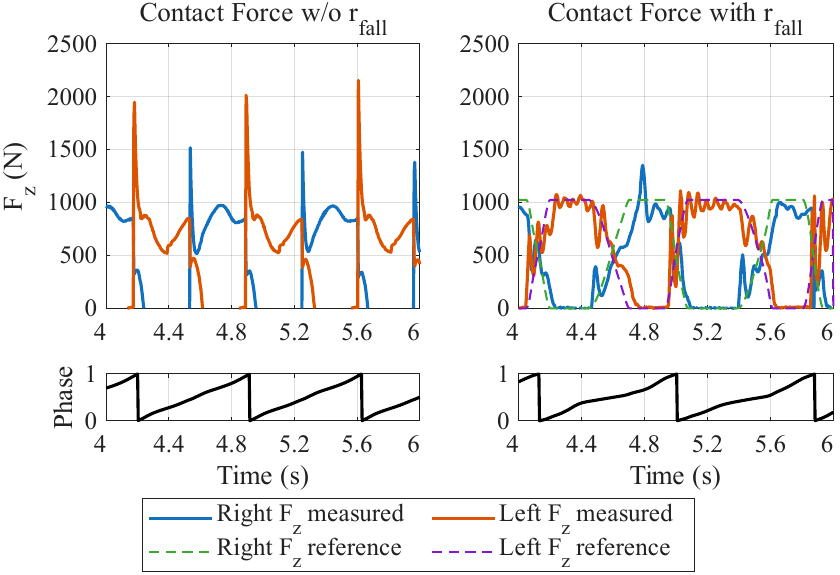}
    \caption{\textbf{Effect of the fallibility rewards.} The reward mitigates early-contact impacts under motor failures and reduces impulsive forces that can reach up to 2000 N on the 100 kg robot TOCABI in real-world experiments.}
    \label{fig:force_ref_phase}
\end{figure}

\begin{figure*}[t]
    \centering
    \includegraphics[width=1\linewidth]{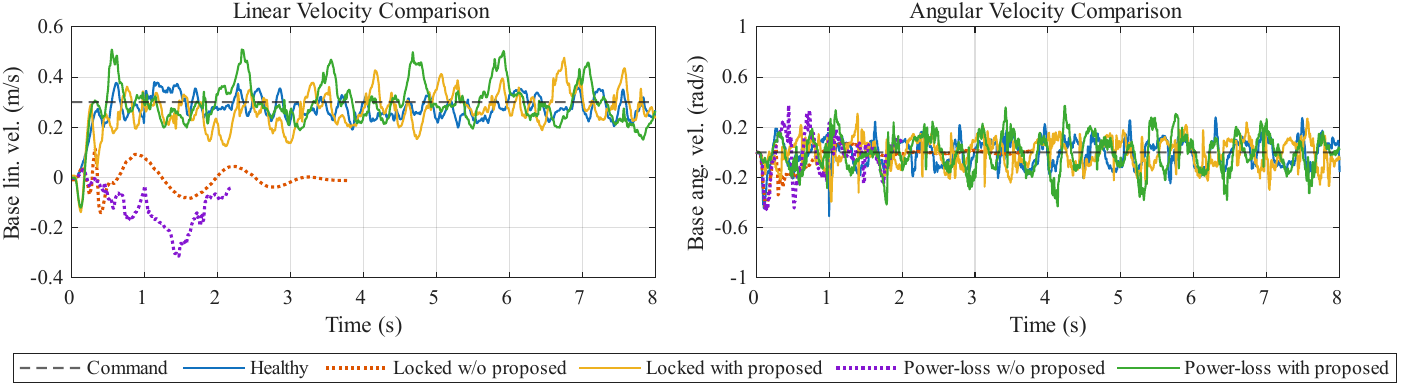}
    \caption{\textbf{Comparison of linear and angular velocity tracking performance in real-world experiments under different motor failure scenarios.} The plots show base linear velocity (left) and base angular velocity (right) over time for the commanded velocity, healthy case, and motor failure cases (locked joint, power loss) with and without the proposed method. The results demonstrate that the proposed approach maintains stable velocity tracking even under motor failures.}
    \label{fig:linear_angular_twocolumn}
\end{figure*}

\subsection{Ablation Study}
\noindent This section evaluates TOLEBI's key components: Joint status observation, fallibility rewards, phase modulation and curriculum learning. 
Table~\ref{tab:ablation_study_with_mbe_compact} shows the performance metrics Root Mean Square Error (RMSE) and Mean Bias Error (MBE) for both linear and angular velocity tracking averaged over motor failure scenarios.
Removing the joint status observation led to a remarkable performance degradation with particularly low performance under healthy conditions. This indicates that estimating the joint status online is essential for accurate and fault-tolerant locomotion control. Excluding the fallibility rewards also reduced robustness under motor failures as the policy failed to effectively promote safe behaviors, where the contact force reward specifically mitigates early-contact impacts that otherwise destabilize locomotion as shown in Fig.~\ref{fig:force_ref_phase}. 
Similarly, removing the phase modulation resulted in the highest MBE among all ablation settings since the absence of phase modulation failed to shorten the stance duration of the impaired leg. This eliminated the controller's ability to adapt the gait timing. Furthermore, training in the absence of curriculum learning prevented the policy from learning nominal locomotion that resulted in gaits where the robot failed to properly lift its feet even under healthy conditions. The complete model of TOLEBI with all components achieved the lowest tracking errors that highlights their benefits for robust fault-tolerant biped locomotion.





\subsection{Sim-to-Real Validation Experiments}
\begin{figure}[b!]
    \centering
    \includegraphics[width=1\linewidth]{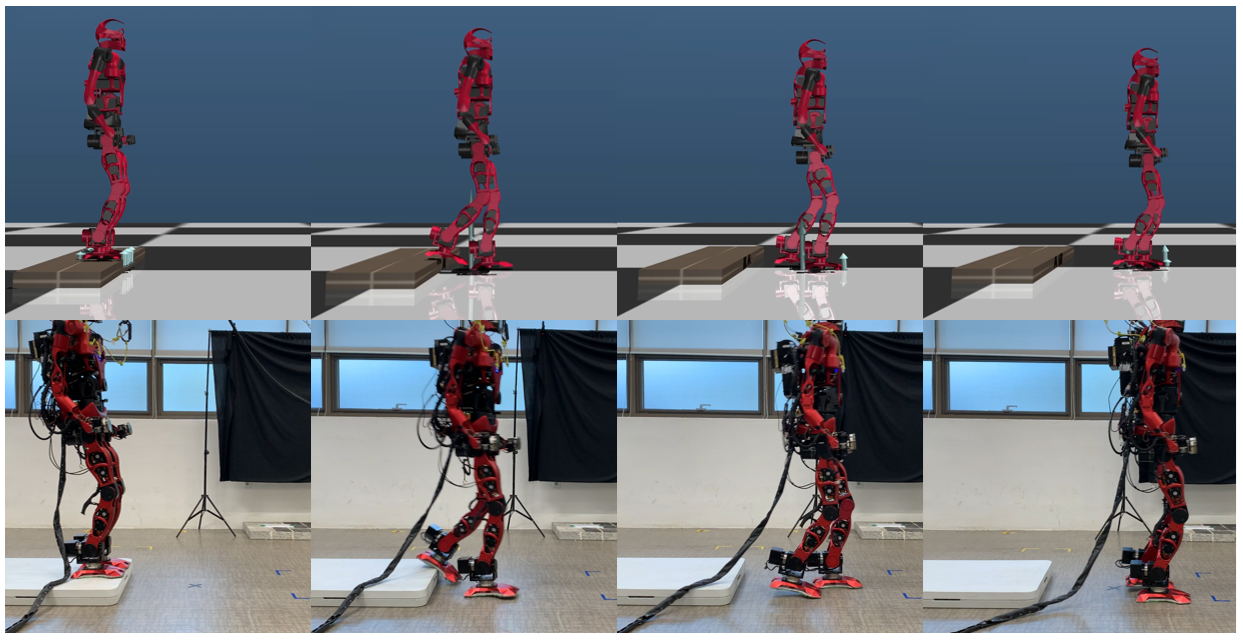}
    \caption{\textbf{Validation of stair descent under motor failure conditions.} TOLEBI enables the humanoid robot TOCABI to successfully perform stair descent in both MuJoCo simulation and real-world experiments.}
    \label{fig:stairdescent}
\end{figure}
\subsubsection{Walking Straight Experiments}
To evaluate the robustness of TOLEBI in real-world settings, flat-ground walking straight experiments were conducted under both healthy and motor failure conditions. As shown in Fig.~\ref{fig:linear_angular_twocolumn}, the robot was commanded to walk straight with a forward velocity of $v_x = 0.3m/s, v_y=0, w_z=0$. The comparison between simulation and real-world trials demonstrates that TOLEBI maintains stable in linear and angular velocity tracking under joint locking and power loss scenarios. These demonstrate TOLEBI's capability to handle faults. 

\subsubsection{Stair Descent Experiments}
In addition to flat-ground walking, stair descent tasks with 9 $cm$ steps were conducted to further evaluate the robustness of TOLEBI under motor failures. Stair environments expose humanoid robots to frequent risks of external impacts or internal circuit faults, making motor failures more likely during descent. To assess TOLEBI’s ability to handle such challenging scenarios, the humanoid robot TOCABI was tested in both MuJoCo simulation and real-world experiments. As illustrated in Fig.~\ref{fig:stairdescent}, the robot successfully descended stairs in the presence of joint locking or power loss conditions. Notably, no additional terrain-specific curriculum learning for stair descending was employed, yet the policy generalized its learned skills to this unseen task. These results demonstrate TOLEBI’s robustness in maintaining balance and stable locomotion despite unexpected motor failures, confirming its practical applicability for fault-tolerant humanoid locomotion.

\section{Conclusion} \label{sec:conclusion}
\noindent This paper presented TOLEBI, a fault-tolerant reinforcement learning framework for biped locomotion. TOLEBI integrates joint masking for diverse failure scenarios, an estimator for joint status inference, and a curriculum learning strategy that gradually exposes the policy to nominal walking, motor failures, and external disturbances. Demonstrating successful sim-to-real transfer, the learned policy enabled the humanoid robot TOCABI to maintain stable locomotion under motor failures as joint locking and power loss. Moreover, the learned policy achieved straightforward walking and stair descent without additional terrain-specific training. Future work will focus on handling multiple simultaneous failures and robust locomotion in unstructured environments toward resilient locomotion strategies.

\section*{Appendix}
\subsection{Randomization Parameters}

\begin{table}[h!]
    \centering
    \caption{Randomization parameters.}
    \label{tab:randomization_params_compact}
    \begin{tabular}{lll}
        \toprule
        & \textbf{Parameter} & \textbf{Randomization Range} \\
        \midrule
        \multirow{7}{*}{\rot{\textbf{Domain Randomization}}}
        & Command Lin. Velocity & $v_x \in U[-0.3, 0.6]$ $m/s$, \\
        & & $v_y \in U[-0.3, 0.3]$ $m/s$ \\
        & Command Ang. Velocity & $\omega_z \in U[-0.5, 0.5]$ $rad/s$\\
        & Push Perturbation & Force $F \in U[50, 250]$ $N$, \\
        &  & Time $T \in U[0.1, 1]$ $s$\\
        & Base Velocity Noise & Lin vel $v \in U[-0.025, 0.025]$ $m/s$, \\
        &  & Ang vel $w \in U[-0.02, 0.02]$ $rad/s$ \\
        \midrule
        \multirow{7}{*}{\rot{\textbf{Dynamics Randomization}}}
        & Link Mass & $U[0.6, 1.4]$ $kg$ \\
        & Link Inertia & $U[0.6, 1.4]$ $kg*m^2$ \\
        & Link Center of mass & $U[0.6, 1.4]$ $m$ \\
        & Motor Constants & $U[0.9, 1.1]$ \\
        & Joint Friction & $U[0.6, 1.4]$ $Nm$ \\
        & Joint Damping & $U[0.6, 1.4]$ $Nm*s/rad$ \\
        & Actuation Delay & $U[0.5, 1.5]$ $ms$ \\
        \bottomrule
    \end{tabular}
\end{table}



\bibliographystyle{IEEEtran}

\bibliography{references}





\end{document}